\pdfoutput=1
\documentclass[conference]{IEEEtran}
\IEEEoverridecommandlockouts
\usepackage{cite}
\usepackage{amsmath,amssymb,amsfonts}
\usepackage{algorithmic}
\usepackage{graphicx}
\usepackage{CJKutf8}
\usepackage{xcolor}
\usepackage{threeparttable}
\usepackage{textcomp}
\usepackage{stfloats}
\PassOptionsToPackage{hyphens}{url}
\usepackage[hidelinks]{hyperref}   
\usepackage{xurl}               
\hypersetup{
colorlinks=false,
pdfborder={0 0 0},
breaklinks=true
}

\def\BibTeX{{\rm B\kern-.05em{\sc i\kern-.025em b}\kern-.08em
    T\kern-.1667em\lower.7ex\hbox{E}\kern-.125emX}}
\begin{document}

\title{Unjamming Deferrals: Supporting Students Through JAMN Bot, a Mobile RAG-Grounded Policy Assistant}

\author{
\IEEEauthorblockN{
Mark Farinas*,
Alesandra Garcia*,
Maria Reyes*,
Jethro Nueva*,
Nawaf Qureshi*,\\
Mahmoud Elsaadany,
and Shokry Shamseldin
}
\IEEEauthorblockA{
\textit{Department of Computer Science} \\
\textit{MacEwan University}\\
Edmonton, Canada \\
\{farinasm, garciaa37, reyesm26, nuevaj, qureshin4\}@mymacewan.ca \\
\{elsaadanym, shamseldins\}@macewan.ca
}
\thanks{\textsuperscript{*}These authors contributed equally to this work.}
\thanks{This is the extended version of a paper published in the proceedings of the 2026 IEEE Smart World Congress (SWC), pp.~810--817, as part of IEEE SmartEdu 2026 (DOI: \href{https://doi.org/10.1109/SWC69865.2026.11763986}{10.1109/SWC69865.2026.11763986}). It restores material that was removed from the conference version to meet the page limit, including the temperature sensitivity analysis, the interface rationale, and the full limitations discussion. Please cite the published conference version.}
}

\maketitle
\begin{abstract}
In some educational institutions, workflows such as the exam deferral process can be slow, inconvenient, and inconsistent, which creates the need to implement novel digital systems. This paper presents Schedulater, a mobile app for managing university exam deferrals. JAMN Bot is a Large Language Model (LLM)-powered chatbot integrated into Schedulater that provides context-aware support to help users understand the deferral process. The system features a Retrieval-Augmented Generation (RAG) pipeline using Ollama (Llama 3.2 3B) for self-hosted inference which enables policy-grounded responses without sacrificing privacy. We evaluate the system using a 43-item benchmark spanning Exam Deferral, Privacy, Scheduling, Cross-Policy, and Adversarial \& Out-of-Scope questions, and report inter-rater agreement among our human evaluators alongside retrieval hit-rate metrics. Subsequently, we utilize an LLM-as-a-judge approach with GPT-5.4, revealing slight-to-moderate alignment with our raters that improves with answer quality, and check the ranking against a panel including Gemini 3.5 Flash and Qwen3.5 397B. The results of our study display trade-offs between open-weight and frontier models, as well as LLM-human evaluation alignment with answer quality. Schedulater demonstrates the feasibility of integrating lightweight LLMs into university administrative workflows with structured retrieval. 
\end{abstract}

\begin{IEEEkeywords}
Educational Chatbot, Mobile Applications, Large Language Models (LLMs), Retrieval-Augmented Generation (RAG), Institutional Policy Navigation
\end{IEEEkeywords}

\section{Introduction}

\subsection{Motivation}
Universities heavily depend on digital systems to coordinate academic operations, yet students often struggle with navigating institutional policies. Matters such as deferred examinations and scheduling procedures are often difficult to interpret without guidance. Conventional support channels, including email correspondence and in-person inquiries, can be slow and may yield inconsistent responses, leading to student uncertainty~\cite{Nadir2025}. These barriers to understanding university administrative processes reveal a clear demand for a user-friendly and reliable system that can provide timely assistance to users~\cite{Nadir2025}.

We developed Schedulater, an exam deferral mobile app, as a solution to streamline university deferral processes. The app supports three types of users: student, faculty, and admin. Students message their professors to request deferrals and gain access to submit a deferral request. Faculty can approve or deny student requests and can also volunteer to proctor deferred exams. Admins are in charge of booking rooms for exam deferrals and assigning proctors to monitor those exam sessions.

One of Schedulater's key features is JAMN Bot, a Large Language Model (LLM)-powered conversational assistant integrated within the mobile app. The chatbot uses an open-source LLM self-hosted through Ollama\cite{ollama,Marcondes2025} as the core service for JAMN Bot, allowing it to generate responses for users. JAMN Bot acts as a policy assistant by answering student queries and guiding users through processes such as submitting deferral requests and day-of-exam procedures. JAMN Bot’s current implementation focuses on the student experience and guiding them through any confusion that arises in the deferral process.

\subsection{Existing Gaps}
Many universities have developed personalized chatbots; however, current systems are more focused on tutoring, engagement, or broad LLM access, which leaves a gap in policy navigation support. The University of Michigan utilizes two different AI-powered tools: Go Blue and Maizey. Go Blue is an AI assistant for campus life that provides real-time information for campus services, including dining and transit\cite{umichGoBlue}, which allows users to upload their own data and documents for a personalized experience. Faculty use the second tool, Maizey, to facilitate tutoring for their courses. Maizey uses Retrieval-Augmented Generation (RAG) to ensure responses come directly from the uploaded materials. The University of California, Irvine launched ZotGPT ClassChat, a tool where instructors upload course materials so the chatbot can provide students with in-depth explanations to their questions\cite{ucirvineZotGPT}. Arizona State University launched the CreateAI platform, providing students, faculty, and staff access to over 50 generative AI models to which they can add instructions and system prompts for customization\cite{asuCreateAI}. They also have a chatbot, Sunny, that assists students with course navigation and scheduling tasks\cite{ascode,edtech_chatbots2025}.

JAMN Bot is different from the aforementioned chatbots in that it is preconfigured with policies and integrated into the workflow, to provide students with immediate responses grounded in institutional policy documents. It combines academic workflows with AI for students to receive guidance without delays.
 
\subsection{Contributions}
The contributions of this work include the design and implementation of an LLM-based chatbot integrated in an exam deferral scheduling system within a university setting, along with a case study evaluating its effectiveness as a policy assistant. From this case study, we report three findings: our raters agree strongly on grounding but less so on correctness, model scale does not predict answer quality under a grounding-strict rubric, and the serving stack affects LLM-human agreement on correctness but not on grounding.

\section{Related Work}
\subsection{AI and Mobile-Based Learning Systems}
Due to recent advances in AI-driven educational tools, the ecosystem of mobile-based assistants utilized in academic and administrative tasks continues to grow\cite{Chen2020,duolingo}. Two popular examples of these learning applications are the platforms Duolingo and Quizlet, both of which demonstrate how AI can be embedded into everyday workflows for many people broadly in the education space\cite{duolingo,quizlet}. These systems, which provide adaptive explanations, practice support, and personalized feedback, highlight the effectiveness of AI tools on their users due to the improved accessibility and engagement for these mobile-based learning environments\cite{Chen2020}.

\subsection{LLM-Powered Assistants and Chatbots in Education}
Historically, the early adoption of AI chatbots in educational institutions focused primarily on answering common questions for students\cite{Page2017}. However, only more recently have they been developed to utilize LLMs, which are more capable of handling open-ended questions from students\cite{Chen2020}. Aside from pure capability improvements, these LLM-based systems have enhanced user experience when prompted to explain concepts, summarize material, generate practice questions and subsequent answers from course content by supporting more natural interactions~\cite{kabir2023llm,Jhajj2025,Davar2025,hicke2023}. Despite limited integration into institutional workflows, there is a growing demand for and a shift towards more conversational educational support systems.
 
\subsection{RAG in Educational Systems}
RAG is a key technique that grounds responses in external knowledge sources to improve the factual reliability of language model outputs\cite{ragpaper}. In these systems, user queries retrieve the relevant context from a collection of structured documents, which are then provided to the model during generation\cite{ragpaper}. This approach reduces hallucinations and improves factual consistency compared to standalone LLMs~\cite{shuster2021}.

In the context of educational environments, RAG has been applied to question-solving systems, tutoring support, and other academic assistance tools. Using RAG in LLM-powered tools enables the models to generate explanations based on lecture notes, textbooks, or institutional documentation\cite{Chen2020}. These properties make RAG particularly suitable for institutional applications where accuracy and traceability are critical.

\subsection{Small Models vs Frontier Models in Deployment}
JAMN Bot is powered by Llama 3.2 3B, a relatively small language model~\cite{llama3}. Smaller models are easier to deploy and can be better suited for privacy-sensitive environments, such as universities, while being more efficient than large frontier models\cite{LinJhajj2024,Ollion2024}. Additionally, behavioural predictability and overhead reduction are maintained with RAG\cite{ragpaper}. Notably, there will always be trade-offs between smaller local models and large frontier models\cite{llama3}. Large models are capable of superior reasoning and broader world knowledge; however, they are also more costly, rely on external cloud APIs and providers which can raise privacy concerns, and require greater computational resources\cite{GPT4}. Institutions will also have to consider that frontier LLMs used via APIs are still subject to versioning and deprecation which impact their stability over time\cite{Ollion2024}. 

Ultimately, despite the limitations in capability, locally-served open-weight models offer adequate performance and allow behaviour verification due to fixed model files\cite{LinJhajj2024}. Although larger models such as Kimi K3~\cite{kimiteam2026kimik3openfrontier}, GLM 5~\cite{glm5}, and Deepseek V4~\cite{deepseekv4} have become easier to serve due to advances such as vLLM~\cite{vLLM}, we consider them out of reach for most institutions. For on-device deployment, hardware constraints pose an even greater hurdle.

\subsection{LLM-as-a-Judge Methodology and Evaluation Practice}
Recent work in LLM evaluation explores the use of LLMs as automated evaluators of other models' outputs, commonly referred to as LLM-as-a-judge frameworks\cite{Zheng2023}. This approach scores outputs on criteria such as correctness, relevance, and helpfulness. However, it has been shown in previous studies that LLM-as-a-judge evaluations encounter calibration issues, especially for subjective criteria like helpfulness and quality of reasoning, where an LLM's judgements can systematically diverge from human annotations\cite{Zheng2023}. A notable issue with using a single LLM judge to evaluate multiple models is that the judge may favour its own answers. Following prior work\cite{verga2024}, we score with a panel of independent judges and report the results in Section~\ref{sec:jury}.

\section{System Design}
\begin{figure*}[b]
    \centering
    \includegraphics[width=\linewidth]{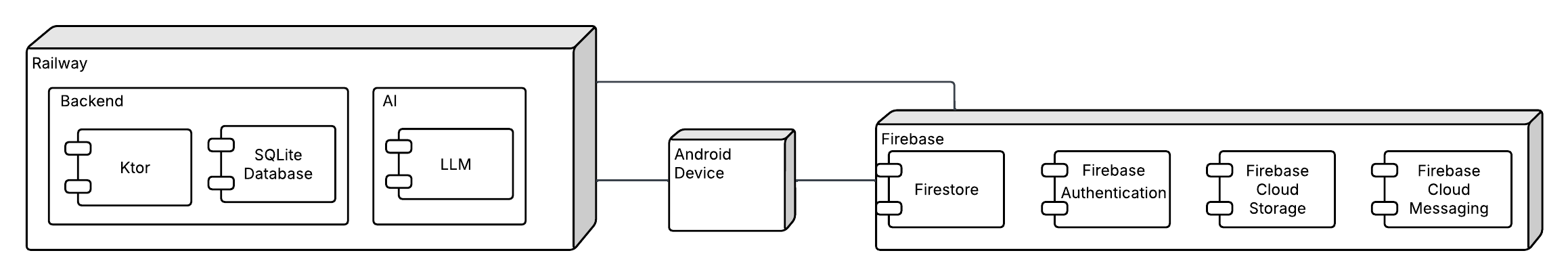}
    \caption{Schedulater's deployment architecture: Communication between the Android client, Railway (hosting the Ktor backend, SQLite store, Ollama-served LLM), and Firebase (for authentication, document storage, cloud storage, and FCM notification)}
    \label{fig:deployment}
\end{figure*}

\subsection{Deployment Context: the Schedulater Platform}
Schedulater is a mobile platform designed to support exam deferrals in university settings. It connects students, instructors, and administrators within a single workflow for requesting, reviewing, and scheduling deferred examinations, while providing status updates throughout the process.
The app uses a Jetpack Compose Android client and a Kotlin backend implemented with Ktor, and deployed on Railway. Firebase is used for authentication, push notifications (FCM), and document storage, while SQLite serves as the embedded vector store for retrieval operations (Fig.~\ref{fig:deployment}). The backend handles core functionality including deferral requests, messaging, room scheduling, and coordination between system components, including the JAMN Bot policy assistant.
JAMN Bot is built into Schedulater, allowing students to ask questions related to the deferral process. When a query is submitted, it is sent to the backend, which manages the AI pipeline. It includes retrieving relevant knowledge, constructing the prompt, and forwarding the prompt to the self-hosted language model served via Ollama. The response is then returned to the client and displayed on the interface. Deploying the backend and LLM separately allows the model to run locally or as a dedicated service, enabling easier independent model updates that improve cost, privacy, and performance.

\subsection{UI Rationale}
\begin{figure}[htbp]
    \centering
    \includegraphics[width=\linewidth]{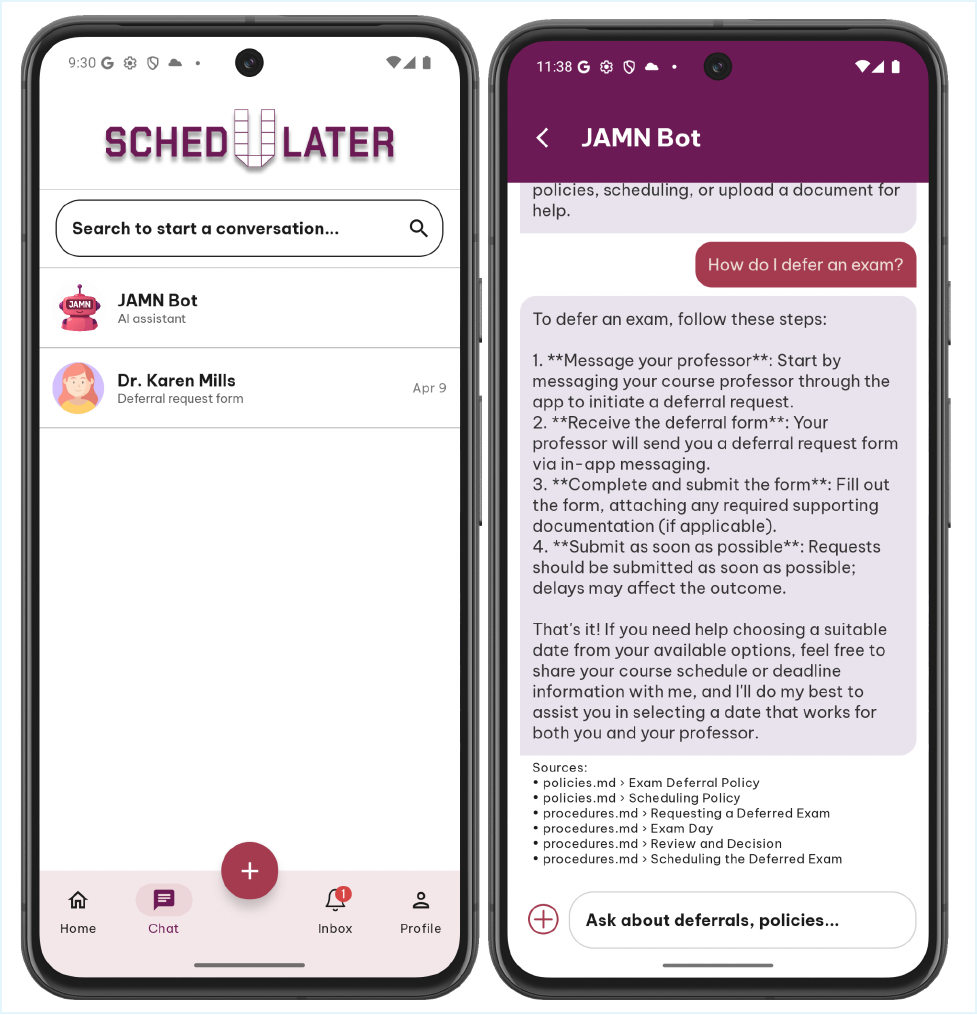}
    \caption{Schedulater's JAMN Bot UI Design integrated with the messaging page (left) and an example of a chat with the AI (right)}
    \label{fig:ui_design}
\end{figure}

JAMN Bot is embedded directly into the student interface and appears at the top of the messaging tab for easy access. This design choice is based on the concept that in-context assistance reduces task-switching and increases the likelihood of usage at the point where students experience uncertainty\cite{rogersmonsell1995}.
The JAMN Bot interface provides a conversational experience with a message history and citations that display the source document and the section of the retrieved context to provide improved transparency, as shown in Fig.~\ref{fig:ui_design}. This allows users to verify the grounding of each answer and reduces over-reliance on model-generated content. The system enforces explicit refusal behaviour. When a query is blocked by guardrails or when retrieval yields no relevant information, the assistant returns a standardized refusal message without fabricated sources. JAMN Bot is designed to avoid speculative or unsupported answers by prioritizing correctness and institutional trustworthiness over completeness.

\section{RAG Pipeline}
\subsection{Knowledge Base}
To effectively answer questions regarding policies, privacy, and scheduling, JAMN Bot utilizes a RAG pipeline. Pre-trained language models have promising capabilities; however, they are also prone to hallucinations\cite{ragpaper}. Our LLM-powered chatbot ensures that outputs remain consistent by grounding its responses in Markdown documents describing MacEwan University policies\footnote{Loosely inspired by MacEwan University's Deferral Policy: \url{https://www.macewan.ca/data/policies/policies/academics/student_deferrals_procedure/student_deferrals_procedure.pdf}} and the procedural workflow of Schedulater.

\subsection{Retrieval and Indexing}
Markdown documents are split into chunks of about 1500 characters with a 100-character overlap during ingestion. This is done using a header-aware chunking strategy. Each chunk is embedded using the nomic-embed-text model\cite{nussbaum2025nomic} via Ollama and stored in a local SQLite database. At query time, the user’s question is embedded and compared against stored chunk vectors using cosine similarity computed in Kotlin. The retriever filters results using a similarity threshold of 0.3 and selects the top-6 most relevant chunks to construct the context. In practice the threshold never binds at this corpus size, as the lowest similarity we observed was 0.505, so every question retrieves six chunks.

\subsection{Prompt Construction and Generation}
The prompt is structured into three parts consisting of system instructions, the user query, and the retrieved context. The system instructions tell the model to  only use information from the Schedulater knowledge base and to avoid unsupported or speculative responses. Responses are generated using our production-level LLM at temperature = 0.3, configurable through environment variables, keeping variation across evaluation runs controlled.

\subsection{Guardrails and Reliability}
As previously mentioned, a layered guardrail system is implemented in the app for security. Prior to searching for an answer, JAMN Bot checks for malicious use, such as prompting to reveal sensitive data. If triggered, the query is blocked immediately. When a retrieval does not return sufficiently relevant context, the system produces a controlled refusal response instead of generating an answer. This ensures that the model does not hallucinate unsupported information when context is missing. Regex serves only as a surface level first-pass check and may miss paraphrased inputs or prompt-injection attacks. Therefore, robustness comes from layering the regex preflight with retrieval validation and a controlled refusal mechanism, rather than from any single check\cite{liu2024formalizing,Bastani25}. This layered approach could be partially extended using LLMs\cite{perez-etal-2022-red} to help reduce the human costs of iterating on and improving the safety and security of our system.

\section{Evaluation}

\subsection{Setup}

The evaluation is based on a 43-item benchmark dataset spanning five operational categories relevant to the Schedulater system: Exam Deferral (10), Privacy (5), Scheduling (5), Cross-Policy (15), and Adversarial / Out-of-Scope (8). Each question is paired with a manually written ground-truth answer. The dataset is designed to evaluate both factual retrieval and workflow reasoning within a university policy context. The benchmark includes a structured adversarial split of Hard Adversarial prompts (items 5, 34, 36, 39) as well as Soft Adversarial / Out-of-Scope (OOS) prompts (items 37, 40–43).

Scoring was performed at two distinct points in this study: first, to evaluate the production deployment of JAMN Bot (Llama 3.2 3B  Q4\_K\_M via Ollama on Railway) against human raters and an LLM judge, and second, to evaluate the performance gap between models of different sizes via OpenRouter\cite{openrouter}. In both scoring instances, the AI-generated answers in response to the benchmark questions were scored against the same rubric on a 0-2 scale on Groundedness (G) and Correctness (C). Groundedness measures whether the response is supported by the knowledge base, while Correctness measures whether the responses are accurate and factually correct.

The inter-rating consensus was taken as the average of three raters (R1, R2, R3). For Cohen's linear-weighted kappa\cite{cohen1968}, which takes an ordinal category as its input, the consensus was taken as the median of the three raters rounded to the nearest integer. The LLM judge, GPT-5.4 via OpenRouter, was given a system prompt with three calibration examples, which were high (G=2, C=2), mid (G=1, C=1), and low (G=0, C=0). These examples were used to aid in aligning its evaluations with the human rubric. All judging was performed at temperature 0 so that the scoring is as close to deterministic as the hosted API allows, and so the variance we observe comes from the answers generated rather than the evaluation\cite{llmJudge}.

For the multi-model comparisons, seven models were evaluated using the same 43-item benchmark and precomputed retrieved chunks for each question to ensure fairness. The retrieval and prompting stages were re-implemented in Python for these runs, using the same chunking parameters, embedding model, similarity threshold, and preflight patterns as the deployed Kotlin backend, so that every model received identical context. The evaluated models included frontier models such as GPT-5.4\cite{gpt5} and Opus 4.5\cite{anthropic2025opus45}, as well as open-weight models such as Llama 3.2 and 3.3\cite{llama3}, Command R7B\cite{coherer7b}, and DeepSeek V4 Pro\cite{deepseekv4}. These models were chosen in particular to reflect possible deployment trade-offs between model size and performance. Across both tests, whether it be the intra-architecture serving comparison or the inter-model capability comparison, generation temperature was held constant at 0.3 across every model on every run in triplicate. Every query logs the retrieved chunks, similarity scores, latency, and a final response to be analyzed in the benchmark pipeline.

\subsection{Manual Scoring of Benchmarks}

\begin{table}[htbp]
\caption{Section-Level Human-Scored Performance: Llama 3.2 3B Q4\_K\_M on Ollama via Railway}
\label{tab:section_eval}
\centering
\scriptsize
\resizebox{\columnwidth}{!}{%
\begin{tabular}{|l|c|c|l|}
\hline
\textbf{Sections} & \textbf{Groundedness} & \textbf{Correctness} & \textbf{Notes} \\
\hline
Sec 1: Exam Deferral (10) & 2.000 & 1.367 & Best-performing category \\
Sec 2: Privacy (5) & 1.800 & 1.333 & Moderate consistency \\
Sec 3: Scheduling (5) & 0.867 & 0.733 & Weakest category \\
Sec 4: Cross-Policy Reasoning (15) & 1.714 & 1.311 & Weak synthesis behaviour \\
Sec 5: Adversarial \& OOS (8) & 1.389 & 1.208 & Soft-adversarial skewed \\
\hline
{Overall (43)} & {1.632} & {1.240} & Combined $(G+C)/2 = 1.436$ \\
\hline
\end{tabular}%
}
\end{table}

The data in Table~\ref{tab:section_eval} elucidates how the production-level model performs against the 43-item benchmark. As seen in Table~\ref{tab:section_eval}, Llama 3.2 3B Q4\_K\_M performs better on Groundedness (1.632) compared to Correctness (1.240), indicating that the model is more reliable with respect to the knowledge base as opposed to answering complex questions. In addition, the combined score (1.436) signifies the solid foundation of JAMN Bot and its ability to improve.

Based on data from Table~\ref{tab:section_eval}, the Scheduling Policy performs the weakest regarding human evaluation (G = 0.867, C = 0.733), while the Exam Deferral Policy displays the strongest result (G = 2.000, C = 1.367). Additionally, the Adversarial/Out-of-Scope section is uniformly difficult (G = 1.389, C = 1.208) while the preflight regex successfully blocks all four of the hard injection prompts (items 5, 34, 36, 39).

\subsection{Inter-Rater Agreement}
\label{sec:irr}
Before comparing the human consensus against the LLM judge, we will establish how consistently the three raters agreed with each other. We report Krippendorff's ordinal $\alpha$\cite{krippendorff2004} alongside Fleiss' $\kappa$\cite{fleiss1971}, as a weighted ordinal measure is better suited to a multi-rater 0-2 scale than an unweighted measure\cite{antoine2014}. Groundedness is reported over 39 items, as the four Hard Adversarial items (5, 34, 36, 39) are blocked by the preflight regex before retrieval and are left undefined. Correctness is reported for all 43 items.

\begin{table}[htbp]
\caption{Inter-Rater Agreement Among the Three Human Raters}
\label{tab:interrater}
\centering
\scriptsize
\begin{tabular}{|l|c|c|l|}
\hline
\textbf{Dimension} & \boldmath$\alpha$ & \textbf{Fleiss'} \boldmath$\kappa$ & \textbf{Interpretation} \\
 & \textbf{(ordinal)} & & \\
\hline
Groundedness ($n$=39) & 0.894 & 0.848 & Almost perfect \\
Correctness ($n$=43) & 0.687 & 0.519 & Substantial / moderate \\
\hline
\multicolumn{4}{l}{$^{\mathrm{a}}$$\alpha$ = Krippendorff's ordinal alpha; pairwise linear-weighted} \\
\multicolumn{4}{l}{Cohen's $\kappa$ averaged 0.87 (Groundedness) and 0.60 (Correctness).} \\
\multicolumn{4}{l}{$\kappa$ is averaged pairwise because weighted $\kappa$ does not extend} \\
\multicolumn{4}{l}{to more than two raters\cite{antoine2014}.} \\
\end{tabular}
\end{table}

As seen in Table~\ref{tab:interrater}, agreement between the raters is high for Groundedness ($\alpha$ = 0.894) with 89.7\% of the items scored unanimously. Correctness is lower ($\alpha$ = 0.687) with 55.8\% of the items scored unanimously. Fleiss' $\kappa$ reads lower than $\alpha$ on both dimensions because it treats the scores as nominal categories and gives no partial credit when raters land on adjacent scores, which happens often on Correctness where raters split between 1 and 2. The per-rater means were closely clustered for Groundedness (1.641, 1.615, 1.641), while one rater scored Correctness slightly stricter than the other two (1.140, 1.279, 1.302). Notably, the strong agreement between our raters indicates that the human consensus is a stable reference to compare the LLM judge against.

\subsection{Large Language Model (LLM) Judge Evaluation}

To complement human evaluation, an automated evaluation pipeline was implemented using GPT-5.4 as an LLM judge applying the same 0-2 scoring rubric, with three calibration examples supplied in the judge prompt. The judge was given the retrieved chunk text as its context so that it assesses Groundedness against the same material the model was shown.

\begin{table}[htbp]
\caption{LLM Judge vs.\ Human Evaluation with Inter-Rater Agreement}
\label{tab:judge_agreement}
\centering
\scriptsize
\resizebox{\columnwidth}{!}{%
\begin{tabular}{|l|c|c|c|l|}
\hline
\textbf{Metric} & \textbf{Human} & \textbf{LLM Judge} & \boldmath$\kappa$ & \textbf{Interpretation} \\
 & \textbf{Avg.} & \textbf{Avg.} & & \\
\hline
Groundedness & 1.632 & 1.667 & 0.400 & Moderate agreement \\
\hline
Correctness & 1.240 & 1.442 & 0.144 & Slight agreement \\
\hline
\multicolumn{5}{l}{$^{\mathrm{a}}$Run 1 of Llama 3.2 3B Q4\_K\_M on Ollama via Railway.} \\
\multicolumn{5}{l}{$^{\mathrm{b}}$$\kappa$ = Cohen's linear-weighted kappa, human consensus} \\
\multicolumn{5}{l}{vs.\ GPT-5.4 judge (temperature 0, chunk text supplied).} \\
\end{tabular}%
}
\end{table}

As shown in Table~\ref{tab:judge_agreement}, the LLM judge and the human raters are closely aligned on Groundedness as the gap between them is only 0.035. The gap in Correctness is larger at 0.202, where the LLM judge scores higher than the human consensus. The judge is willing to award credit for Correctness more readily than our raters are, which aligns with previous literature on LLM judges being permissive on accuracy-style criteria\cite{Zheng2023}.

Cohen’s linear-weighted kappa values are reported in Table~\ref{tab:judge_agreement} ($\kappa_G$ = 0.400, $\kappa_C$ = 0.144). The Landis \& Koch interpretation\cite{landis1977} for Groundedness and Correctness using the kappa values is “moderate” and “slight”, respectively, which gives us a floor for the judge-human agreement on the production-level model.

\subsection{Multi-Model Comparison}

Utilizing seven different models, we ran triplicate benchmarking against the 43-item dataset using the same precomputed retrieved chunks and generation temperature set to 0.3. Subsequently, GPT-5.4 judged every AI-generated output under the same scoring rubric at temperature 0.

\begin{table}[htbp]
\caption{Multi-Model Leaderboard (Ranked by Combined = (G+C)/2, mean $\pm$ SD over 3 runs)}
\label{tab:model_leaderboard}
\centering
\scriptsize
\resizebox{\columnwidth}{!}{%
\begin{tabular}{|c|l|c|c|c|}
\hline
\textbf{Rank} & \textbf{Model} & \textbf{Groundedness} & \textbf{Correctness} & \textbf{Combined} \\
\hline
1 & GPT-5.4 & 1.829 & 1.837 & 1.833 $\pm$ 0.018 \\
2 & Llama 3.3 70B Instruct & 1.812 & 1.721 & 1.766 $\pm$ 0.029 \\
3 & Opus 4.5 & 1.726 & 1.775 & 1.751 $\pm$ 0.039 \\
4 & DeepSeek V4 Pro & 1.675 & 1.597 & 1.636 $\pm$ 0.115 \\
5 & Llama 3.2 3B Instruct & 1.718 & 1.481 & 1.600 $\pm$ 0.041 \\
6 & Command R7B (12-2024) & 1.615 & 1.543 & 1.579 $\pm$ 0.044 \\
7 & Llama 3.2 1B Instruct & 1.385 & 0.961 & 1.173 $\pm$ 0.051 \\
\hline
\end{tabular}%
}
\end{table}

Based on the results in Table~\ref{tab:model_leaderboard}, the full-precision Llama 3.2 3B Instruct ranks 5th out of 7 when ranked by the combined results $\frac{G+C}{2}$. Every model in this table was served at full precision through OpenRouter, so this ranking refers to the full-precision counterpart of the model we deploy and not the deployed configuration itself, which we compare against in Section~\ref{sec:instruct}. While GPT-5.4 performs best among the seven models, the difference between its combined result and Llama 3.2 3B Instruct is 0.233. When comparing that to Llama 3.2 1B Instruct, which placed last, the difference is substantial. This is due to 1B being a smaller Llama model with a lesser capacity to follow instructions and reason over retrieved context. Notably, the temperature ablation in Section~\ref{sec:temp} gives us a run-to-run spread of about 0.08 on this benchmark. Only two of the six gaps between adjacent ranks in Table~\ref{tab:model_leaderboard} are larger than that spread, so the top three models and the middle of the leaderboard are better interpreted as groupings than as a strict ordering.

\begin{table}[htbp]
\caption{Per Section Combined Scores (Ranked by Overall)}
\label{tab:section_combined_scores}
\centering
\begin{threeparttable}
\scriptsize
\resizebox{\columnwidth}{!}{%
\begin{tabular}{|l|c|c|c|c|c|c|}
\hline
\textbf{Model} & \textbf{Sec 1} & \textbf{Sec 2} & \textbf{Sec 3} & \textbf{Sec 4} & \textbf{Sec 5} & \textbf{Overall} \\
\hline
GPT-5.4 & 2.000 & 1.833 & 1.800 & 1.865 & 1.542 & 1.833 \\
Llama 3.3 70B Instruct & 2.000 & 1.567 & 1.734 & 1.773 & 1.576 & 1.766 \\
Opus 4.5 & 1.965 & 1.634 & 1.633 & 1.795 & 1.500 & 1.751 \\
DeepSeek V4 Pro & 1.791 & 1.700 & 1.567 & 1.727 & 1.222 & 1.636 \\
Llama 3.2 3B Instruct & 1.967 & 1.534 & 1.500 & 1.615 & 1.167 & 1.600 \\
Command R7B (12-2024) & 1.724 & 1.400 & 1.367 & 1.681 & 1.424 & 1.579 \\
Llama 3.2 1B Instruct & 1.537 & 1.000 & 1.067 & 1.184 & 0.841 & 1.173 \\
\hline
\end{tabular}%
}
\begin{tablenotes}
\small 
\item Section 1 = Exam Deferral Policy, Section 2 = Privacy Policy, Section 3 = Scheduling Policy, Section 4 = Cross-Policy Reasoning, Section 5 = Adversarial \& OOS
\end{tablenotes}
\end{threeparttable}
\end{table}

Each model has Groundedness and Correctness calculated per section to observe which model performed best in each section. From Table~\ref{tab:section_combined_scores}, it is evident that the Adversarial \& OOS section is the weakest category for most models, indicating that it is a domain-wide challenge for RAG systems. This behaviour is expected, as many Adversarial and OOS prompts do not map to any relevant chunks in the knowledge base, leaving the model without grounded context to generate a correct response regardless of model size or capability. When a policy exists for a refused query, our retrieval results (Section~\ref{sec:retrieval}) show that the retriever does not always surface it, which is a separate failure from the absence of a relevant chunk. The Exam Deferral Policy is consistently the strongest across all models, suggesting it is the most structured and rule-based. Cross-Policy Reasoning also scores well despite requiring the model to pull information from policy.md and procedure.md knowledge files in parallel. As well, the weakness in Scheduling Policy that we saw during the human evaluation (Table~\ref{tab:section_eval}) does not carry over to the other models, which tells us that it is an issue specific to our quantized baseline and not a difficulty with the section itself. When focusing on the Exam Deferral Policy, DeepSeek V4 Pro performs worse than both Opus 4.5 (3rd place) and Llama 3.2 3B Instruct (5th place), yet it still ranks 4th overall. For the Privacy Policy, it performs better than both, which may be why it ranks between the two models overall.

\subsection{Scoring on the full-precision Llama 3.2 3B Instruct}
\label{sec:instruct}

We previously outlined that the Landis \& Koch interpretation of the Cohen’s kappa values are “moderate” ($\kappa_G$ = 0.400) and “slight” ($\kappa_C$ = 0.144) for the benchmarked and judged Llama 3.2 3B Q4\_K\_M (Table~\ref{tab:judge_agreement}). Although $\kappa_C$ demonstrates a misalignment between the LLM judge and human raters, it does not imply that the judge is unreliable. Rather, we need to conduct more benchmarks with the models that generated better answer quality to see if it reduces disagreement between the LLM judge and human ratings. 

We manually scored the benchmarks from Run 1 of Llama 3.2 3B Instruct via OpenRouter with the same temperature and 3-rater protocol, following a rubric on a scale from 0-2. Our raters scored the Instruct configuration at mean G = 1.701 and C = 1.434, compared to G = 1.632 and C = 1.240 for the Q4\_K\_M configuration. Comparing the two configurations item by item, Correctness improved on 20 of the 43 items and worsened on 8, while Groundedness improved on 9 items and worsened on 8.

As we hold both the generation temperature and the judge configuration constant across the two runs, we are able to isolate the effect of the serving stack on the judge-human agreement. However, the two configurations differ in more than one respect. Both run the same instruct-tuned model, but they differ in numerical precision, inference provider, and whether the retrieval and prompting stages ran through the deployed backend or the Python re-implementation. We therefore cannot isolate any one of these as the cause.

We recorded the Instruct model configuration’s Cohen’s linear-weighted kappa values as $\kappa_G$ = 0.389 and $\kappa_C$ = 0.463 when comparing the GPT-5.4 and human scores. Using the Landis \& Koch’s scale once again, these values are interpreted as “fair” and “moderate”, respectively, for Groundedness and Correctness. For Groundedness, the two configurations land in effectively the same place ($\kappa_G$ = 0.400 for Q4\_K\_M and 0.389 for Instruct), so the serving stack does not appear to change how closely the judge and our raters agree on grounding. Additionally, the Instruct configuration’s $\kappa_C$ lifts by 0.319 making it substantially higher than that of the Q4\_K\_M configuration, moving it from “slight” to “moderate” agreement.

\begin{figure}[t]
\centering
\includegraphics[width=\columnwidth]{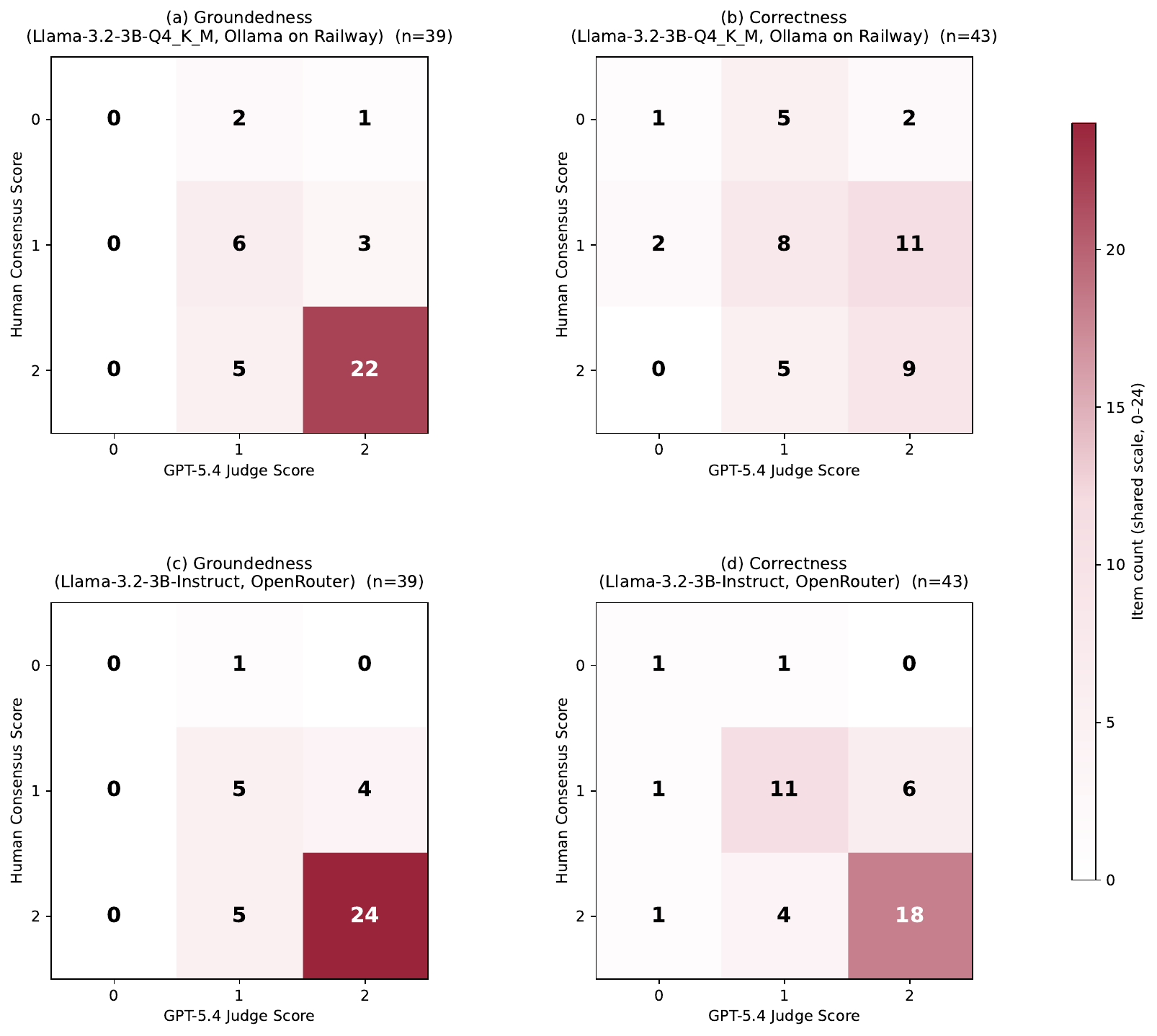}
\caption{Human Consensus vs GPT-5.4 Judge: Confusion Matrices with Llama 3.2 3B Q4\_K\_M on Ollama via Railway (a,b) and Llama 3.2 3B Instruct via OpenRouter (c,d). All panels share a common colour scale.}
\label{fig:confusion_matrices}
\end{figure}

The confusion matrices for the Llama 3.2 3B Instruct model shown in Fig.~\ref{fig:confusion_matrices} demonstrate a dominant diagonal pattern as 29/39 Groundedness items and 30/43 Correctness items sit in perfect LLM judge-human agreement. On the other hand, it is visually evident with Fig.~\ref{fig:confusion_matrices} that the Q4\_K\_M model configuration has many more instances of misalignment in Correctness, as its diagonal reveals only 18/43 items in agreement. The Q4\_K\_M Groundedness diagonal sits at 28/39, which is close to the Instruct configuration and matches what we see in the kappa values. It is important to note that Groundedness is reported as n=39 due to the four Hard Adversarial items (5, 34, 36, 39) flagging the preflight regex prior to retrieval, leaving those four items undefined.

Despite temperature being held constant across both runs, the calculated kappa lift and the apparent improved judge-human scoring agreement cannot be attributed to a single factor. However, we believe the data supports that improved judge-human score agreement on Correctness scales with answer quality, while the agreement on Groundedness holds steady across both configurations.

\subsection{Independent Multi-Judge Panel}
\label{sec:jury}
We utilized GPT-5.4 as both a judge and a candidate in the multi-model comparison, and it ranked itself first on the leaderboard in Table~\ref{tab:model_leaderboard}. To check whether that ranking depends on the judge, we re-scored all seven candidates with two more judges taken from outside the candidate pool, Gemini 3.5 Flash and Qwen3.5 397B. Prior work recommends a panel of judges for this reason\cite{verga2024}. All three judges scored under the same rubric and the same three calibration examples at temperature 0. As well, the judges differ in how much intermediate reasoning they emit by default, so each judge's configuration was held constant across its entire sweep. It is worth noting that all three are proprietary hosted models, so we make no claim of reproducibility or open weights for any of them.

\begin{table}[htbp]
\caption{Multi-Judge Panel: Combined $(G+C)/2$ per Candidate, by Judge}
\label{tab:jury}
\centering
\scriptsize
\resizebox{\columnwidth}{!}{%
\begin{tabular}{|l|c|c|c|}
\hline
\textbf{Candidate} & \textbf{GPT-5.4} & \textbf{Gemini 3.5} & \textbf{Qwen3.5} \\
 & \textbf{(incumbent)} & \textbf{Flash} & \textbf{397B} \\
\hline
GPT-5.4$^{\mathrm{a}}$ & 1.833 & 1.967 & 1.857 \\
Llama 3.3 70B Instruct & 1.766 & 1.932 & 1.819 \\
Opus 4.5 & 1.751 & 1.884 & 1.824 \\
DeepSeek V4 Pro & 1.636 & 1.735 & 1.655 \\
Llama 3.2 3B Instruct & 1.600 & 1.643 & 1.525 \\
Command R7B (12-2024) & 1.579 & 1.719 & 1.582 \\
Llama 3.2 1B Instruct & 1.173 & 0.961 & 0.828 \\
\hline
\multicolumn{4}{l}{$^{\mathrm{a}}$Incumbent judge scoring its own outputs. Rank correlations} \\
\multicolumn{4}{l}{and the self-preference test are reported in the text.} \\
\end{tabular}%
}
\end{table}

All three judges rank GPT-5.4 first and the ordering of the other six candidates is largely preserved, as seen in Table~\ref{tab:jury}. Excluding each judge's own candidate, the Spearman rank correlations are 0.943 between GPT-5.4 and Gemini, 0.886 between GPT-5.4 and Qwen, and 0.964 between the two independent judges. The judges differ in how harshly they score, as Gemini scores higher across the board while Qwen scores lower. Two adjacent pairs change places, Opus 4.5 with Llama 3.3 70B and Llama 3.2 3B with Command R7B, and both pairs are separated by less than the sampling noise measured in Section~\ref{sec:temp}.

To check for self-preference, we compared the difference between GPT-5.4’s score and each independent judge’s score for every candidate. GPT-5.4’s difference on its own outputs is not an outlier against the differences on the other six candidates ($z$ = $-0.51$ against Gemini and $z$ = $-0.44$ against Qwen). We also computed the agreement across the three judges on a matched item set, restricted to the one candidate our raters also scored, which gives $\alpha$ = 0.608 for Groundedness and $\alpha$ = 0.711 for Correctness, close to what our three human raters produced ($\alpha$ = 0.661 and $\alpha$ = 0.643). Notably, GPT-5.4’s margin over the second-ranked model is 0.067, which falls below the 0.08 sampling noise floor. Because of this we are not able to confirm or rule out self-preference at this benchmark size, and we present the panel as evidence that the ranking is robust rather than as a measurement of bias.

\subsection{Retrieval Metrics}
\label{sec:retrieval}
Our benchmarking so far measures the quality of the generated answers, but it does not tell us how well the retriever itself performs. To measure this directly, three human annotators independently labelled the gold chunk or chunks containing the answer for each item. Each annotator saw all ten chunks in a fixed order with no scores, no ranks, and no indication of what the retriever had returned, so the labels are blind to the output being scored. Agreement on these labels was substantial ($\alpha$ = 0.553), and three items were adjudicated as having no supporting chunk in the knowledge base. Because retrieval was precomputed and shared across all seven models, the hit-rate is a property of the retriever and not of any individual model, so we report one set of numbers over the 36 answerable items.

\begin{table}[htbp]
\caption{Retrieval Metrics vs.\ Random-Retriever Baseline ($n=36$)}
\label{tab:retrieval}
\centering
\scriptsize
\begin{tabular}{|l|c|c|l|}
\hline
\textbf{Metric} & \textbf{Retriever} & \textbf{Random}$^{\mathrm{a}}$ & \textbf{Reading} \\
\hline
Hit@1 & \textbf{0.667} & 0.167 & Discriminative \\
Hit@3 & 0.889 & 0.438 & Discriminative \\
Hit@6 & 0.889 & 0.732 & Near-saturated$^{\mathrm{b}}$ \\
MRR & \textbf{0.773} & 0.343 & Discriminative \\
Recall@6 & 0.875 & 0.600 & Near-saturated$^{\mathrm{b}}$ \\
\hline
\multicolumn{4}{l}{$^{\mathrm{a}}$Exact, conditioned on each item's gold-chunk count.} \\
\multicolumn{4}{l}{$^{\mathrm{b}}$Fixed top-6 over a 10-chunk corpus; see text.} \\
\end{tabular}
\end{table}

As seen in Table~\ref{tab:retrieval}, the retriever reaches Hit@1 = 0.667 and MRR = 0.773 against random baselines of 0.167 and 0.343, which tells us that when a relevant chunk exists it usually ends up ranked first. Hit@3 and Hit@6 both reach 0.889 and Recall@6 reaches 0.875. However, our knowledge base only holds ten chunks and the retriever returns a fixed top-6, so a random retriever would already reach 0.732 on Hit@6. We report the set-membership metrics for completeness, but at this corpus size it is Hit@1 and MRR that carry the signal.

The one section where the retriever underperforms is Adversarial \& OOS, where Hit@1 and Hit@6 both sit at 0.500 against a random Hit@6 of 0.692 for that section's gold-count mix. On items 38 and 41 the retriever does not surface the Privacy Policy chunk that governs the request, so the assistant would refuse without the relevant policy in context. We report this as an observation given the small number of items involved.

\subsection{Temperature Sensitivity Analysis}
\label{sec:temp}

\begin{figure}[t]
\centering
\includegraphics[width=\columnwidth]{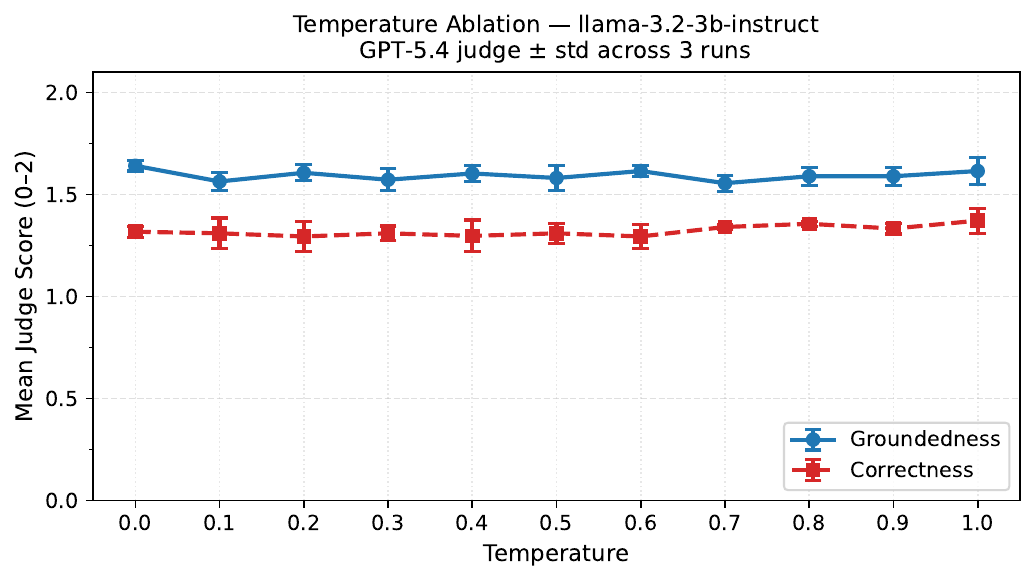}
\caption{Temperature Ablation on Llama 3.2 3B Instruct: mean Groundedness and Correctness across 11 sampling temperatures (0.0 to 1.0).}
\label{fig:temperature_ablation}
\end{figure}

All of our benchmarking was done with a generation temperature of 0.3 to allow for variation in AI-generated answers across seven models in response to the queries, in a non-deterministic manner. By running in triplicate for benchmarking, this deliberate exposure of sample variance allowed us to visualize the variance in model behaviour under realistic conditions. However, we wanted to perform another test on temperature sensitivity using only the Llama 3.2 3B Instruct model to isolate generation temperature as the sole variable. We performed three runs at 0.1 increments in temperature from 0.0 to 1.0. The results show that mean Groundedness has a total spread of 0.085 (1.556 to 1.641) while mean Correctness has a total spread of 0.078 (1.295 to 1.372) across the 11 temperature settings. As the error bars in Figure~\ref{fig:temperature_ablation} show, there is no statistically meaningful trend.

Our data mirrors the null results from Renze and Guven in which their data lacked statistical significance when they tested the performance of five LLMs on multiple-choice problems at different temperatures between 0.0 and 1.0\cite{Renze2024}. We hypothesize that our data follows an analogous mechanism to Renze and Guven’s, as the prompt template explicitly restricts the model from retrieving certain chunks, leading to a lack of deviation from the rubric outcome and preventing temperatures from straying from the dominant trajectory. Notably, however, Renze and Guven mentioned that averaging and statistical noise may undermine genuine temperature-sensitivity results\cite{Renze2024}.

This data allows us to conclude that first, the 0.3 setting used in production sits within a flat region for generation quality, second, there is no clear breakdown point for this model within the tested range, and lastly, the within-model spread of ~0.08 for Groundedness and Correctness can be used as a floor for sampling noise for the leaderboard (Table~\ref{tab:model_leaderboard}). The implications for educational institutions are that, if they choose to adopt a RAG-based system, temperature has minimal impact on outcome variance, and that tuning effort is better directed at prompt structure, model selection, and serving precision, which Table~\ref{tab:model_leaderboard} and Section~\ref{sec:instruct} show move the scores considerably further than sampling temperature does.

\section{Discussion}

\subsection{Effect of the Deployment Configuration}
Educational institutions facing cost or hardware constraints may want to prioritize smaller, cheaper, local models, and will often reach for a quantized model served locally. Our raters recorded a Correctness improvement moving from our deployed configuration to a full-precision hosted one, while the LLM judge recorded a much smaller one and the agreement on Groundedness did not change at all (Section~\ref{sec:instruct}). The two configurations differ in precision, provider, and pipeline implementation, so we report the effect without attributing it to any single one. Notably, this means the judge is least aligned with our raters on exactly the configuration an institution would deploy to save money. The practical implication is that a quantized model's accuracy should be validated against human raters rather than an LLM judge alone, as the automated judge under-reports the cost of quantizing.

\subsection{Raw Capabilities of RAG and Instruction-Following}
Many would expect that larger models would inherently outperform smaller models. However, our findings show that this assumption does not always hold. Observing the multi-model leaderboard comparing the Groundedness and Correctness between all seven models, with a few exceptions, most rankings feel predictable. The frontier models (GPT-5.4 and Opus 4.5) perform well, as expected. However, the rankings of the non-frontier models provide insight into our claim that bigger does not always mean better. Sorting the LLMs by approximate size in descending order, we get: GPT-5.4, Opus 4.5, DeepSeek V4 Pro, Llama 3.3 70B, Command R7B, Llama 3.2 3B, and Llama 3.2 1B, noting that the parameter counts of the closed models are not published. Notably, Table~\ref{tab:model_leaderboard} reveals that the larger DeepSeek V4 Pro model underperforms as it ranks 4th place (G\&C = 1.636), sitting 0.130 below the much smaller Llama 3.3 70B (G\&C = 1.766). That margin is larger than the run-to-run spread of roughly 0.08 we measured, so it is not an artifact of sampling. Despite DeepSeek V4 Pro’s verbose chain-of-thought, its rubric scores are lower than expected\cite{Wu2025}. Due to the rubric's penalization of content that extrapolates beyond the retrieved chunks, models that do not follow the instructions faithfully underperform. 

In general, the RAG benchmark rewards instructional discipline more than raw capabilities. Knowledge of model size is still valuable to infer usefulness to a different degree in the case of Llama 3.2 1B Instruct. Its combined G\&C score of 1.173 can act as a useful performance floor. It is likely that models below this size would be more prone to invalid outputs and hallucinations, as the Instruct configuration implies that the model has already been fine-tuned to improve instruction-following capabilities. A constant systematic pattern appears regardless of model size, revealing that the items from Adversarial \& OOS consistently drop the GPT-5.4 judge scores shown in Table~\ref{tab:section_combined_scores}. While the preflight regex layer cleanly catches all four Hard Adversarial queries, OOS queries, which use paraphrasing or indirect injections, negatively impact the surface-pattern matching which correlates with the G\&C scores- emphasizing the importance of a layered defence system\cite{Bastani25}.
With regard to deployment decisions in educational institutions, practical choices lie between a fine-tuned open-weight model such as Llama 3.2 3B Instruct and the frontier models. However, size-related deployment constraints such as privacy, cost, and latency will play a factor in institutions deciding which model to adopt.

\subsection{Broader implications for AI in Education}

We believe that our work can provide some key insights for the adoption of LLM-based assistants in higher education. Empirical data gathered on Groundedness and Correctness across various models allow us to view the trade-off made between locally deployed smaller open-weight models and frontier models accessed via API calls. Because the 1B model (G\&C = 1.173) is likely insufficient for any meaningful student-facing product, educational institutions must, at minimum, rely on models similar to the 3B and 7B for open-weight models. Although the gap in the scores from Table~\ref{tab:model_leaderboard} indicates that the frontier models have improved performance over the others, the trade-offs would include added cost, latency, and reliance on external APIs which may be constraints at scale. As well, our quantization results show that the choice of model is not the only decision that matters, as the way a model is served affects how reliably its accuracy can be measured. Additionally, we believe that a student-facing AI assistant (even beyond the domain of exam deferrals) will naturally receive queries that may be confusing or adversarial similar to our testing. Because of this, we propose that our layered guardrail pattern (preflight regex, similarity threshold, and empty-retrieval refusal) is a justified design as a first line of defence for chatbots across different domains.

\section{Limitations and Future Work}
JAMN Bot’s knowledge was curated by our team to represent the Schedulater workflow and is not entirely representative of any university’s policies and procedures for deferred exams. To improve the utility of JAMN Bot, future work would include expanding the chatbot’s knowledge base so that it can be helpful for policy beyond exam deferrals. If JAMN Bot were able to provide guidance on academic integrity, grade appeals, major and minor declarations, changing programs, study plans, stress management, or taking a leave of absence, it could become a centralized hub to help students with their academic lives. As well, further collaboration with faculty would help ensure that common student queries can be answered by the chatbot to potentially ease their workloads. Additionally, the chatbot should be enhanced to support the needs of faculty and administrators, as the current implementation is student-focused. However, extending JAMN Bot’s capabilities would likely require a larger LLM. At the same time, JAMN Bot currently relies on simple regex-based guardrails, which are limited in handling more complex queries. As its scope expands, stronger guardrails would be needed to prevent negative outcomes and user experiences\cite{Bastani25}.

To make JAMN Bot as convenient as possible to use, we could perform A/B testing on users to see which placement they prefer for JAMN Bot. As of now, JAMN Bot is the first conversation that shows up when looking at the messaging tab, acting as more of a standalone feature. An alternative to this would be embedding JAMN Bot in the navigation bar.

During manual testing, latency was also recorded, averaging just under one minute per query. To deploy Schedulater into real university environments, we would integrate a vector database to store chunks for more efficient retrieval\cite{Gao2023} and cache common queries to reduce latency. Better retrieval would also address the items in Section~\ref{sec:retrieval} where the governing chunk was never surfaced, a correctness problem and not only an efficiency one. In the future, a more rigorous manual evaluation could be conducted to include a longitudinal study and additional human raters. If more rigorous testing were to be done, we would use a 7-point Likert scale for a more precise rating of Groundedness and Correctness.

Two limitations remain from our own results. Our ten-chunk knowledge base with a fixed top-6 retriever leaves the set-membership retrieval metrics close to their random baselines, and measures such as NDCG would need a substantially larger corpus\cite{Gao2023}. As well, although the judge panel largely preserved the model ranking with no anomalous self-score, GPT-5.4's margin over second place is smaller than the run-to-run variation we measured, so self-preference can neither be confirmed nor ruled out at this benchmark size\cite{verga2024}.

\section{Conclusion}
In this work, we presented JAMN Bot, an LLM-powered policy assistant integrated into Schedulater, an application for university exam deferral workflows. JAMN Bot provides students with context-aware exam deferral guidance via RAG, using a self-hosted LLM served by Ollama. The evaluation of JAMN Bot shows that it performs most effectively on well-structured policy guidance, but still finds Out-of-Scope prompts challenging. Our three raters agreed strongly with one another (Krippendorff's $\alpha$ = 0.894 on Groundedness), giving a stable reference for the LLM judge comparison, and an independent panel of three judges largely preserved the model ranking with no detectable self-preference. Model scale, however, did not predict answer quality under our grounding-strict rubric, as DeepSeek V4 Pro scored 0.130 combined below the much smaller Llama 3.3 70B, a margin wider than the run-to-run spread we measured, because the rubric rewards instruction-following over raw capability. We also found that the serving stack affects judge-human agreement on Correctness but not on Groundedness, which matters for institutions weighing whether to quantize. The results suggest that LLM assistants with well-defined guardrails can reduce confusion in university deferral processes. As well, the differences in performance between closed and open-source models, while present to some degree, inspire confidence for the deployment of open models in educational settings, as the top three models on our leaderboard are separated by less than the run-to-run variation we measured. Overall, institutions considering similar systems should note that JAMN Bot highlights the importance of balancing privacy, accuracy, both monetary and human costs, safety, and the availability of accurate information when introducing an AI chatbot into student administrative workflows. 

\section*{Acknowledgments}
We would like to thank Gaganpreet Jhajj for his expertise and input, API credits and proofreading support. We would also like to thank Yumeko Nomura and Lindsey Searles for proofreading.

\section*{Data and Code Availability}
The benchmark dataset (43 items with ground-truth answers), the knowledge base and derived chunk corpus, the LLM judge prompt with calibration examples, the anonymized per-item human ratings, all model outputs and judge scores for the three judges, the retrieval gold labels, and the analysis scripts are publicly available\footnote{https://github.com/JAMNSoftware/jamn-rag-benchmark} (archived at \cite{zenodolink}). A single script (\texttt{analysis/reproduce.py}) regenerates the paper's headline results from the released files using only NumPy.

\clearpage
\bibliographystyle{IEEEtran}
\bibliography{ref}

\end{document}